\documentclass[conference]{IEEEtran}
\usepackage{cite}
\usepackage{amsmath,amssymb,amsfonts}
\usepackage{algorithmic}
\usepackage{graphicx}
\usepackage{textcomp}
\usepackage[utf8]{inputenc}
\usepackage{xcolor}
\usepackage{subcaption}
\usepackage{booktabs}
\usepackage{hyperref}
\usepackage{multirow}
\usepackage{listings}
\usepackage{cite}

\AtBeginDocument{%
  }

\begin{document}

\title{Enhancing Code Translation in Language Models with Few-Shot Learning via Retrieval-Augmented Generation \\
}

\author{\IEEEauthorblockN{Manish Bhattarai}
\IEEEauthorblockA{\textit{Theoretical Division} \\
\textit{Los Alamos National Laboratory}\\
Los Alamos, NM \\
ceodspspectrum@lanl.gov}
\and
\IEEEauthorblockN{Javier E. Santos}
\IEEEauthorblockA{\textit{Earth \& Environmental Science Division} \\
\textit{Los Alamos National Laboratory}\\
Los Alamos, NM \\
jesantos@lanl.gov
}
\and
\IEEEauthorblockN{Shawn Jones}
\IEEEauthorblockA{\textit{Computer, Computational \& Statistical Sciences} \\
\textit{Los Alamos National Laboratory}\\
Los Alamos, NM \\
smjones@lanl.gov
}
\and
\IEEEauthorblockN{Ayan Biswas}
\IEEEauthorblockA{\textit{Computer, Computational \& Statistical Sciences} \\
\textit{Los Alamos National Laboratory}\\
Los Alamos, NM \\
ayan@lanl.gov}
\and
\IEEEauthorblockN{Boian Alexandrov}
\IEEEauthorblockA{\textit{Theoretical Division} \\
\textit{Los Alamos National Laboratory}\\
Los Alamos, NM \\
boian@lanl.gov}
\and
\IEEEauthorblockN{Daniel O'Malley}
\IEEEauthorblockA{\textit{Earth \& Environmental Science Division} \\
\textit{Los Alamos National Laboratory}\\
Los Alamos, NM \\
omalled@lanl.gov}
}

\maketitle

\begin{abstract}
The advent of large language models (LLMs) has revolutionized the field of code translation, enabling automated translation between programming languages. Despite these advancements, the accuracy and reliability of these models often falter in complex translation tasks due to a lack of contextual understanding. This paper introduces a novel approach to enhance code translation through Few-Shot Learning augmented with retrieval-based techniques. By leveraging a repository of existing code translations, we dynamically retrieve the most relevant examples to guide the model in translating new code segments. Our method, based on Retrieval-Augmented Generation (RAG), significantly improves translation quality by providing contextual examples that the model can learn from in real-time. We chose RAG over traditional fine-tuning methods due to its ability to leverage existing codebases or a locally stored corpus of code, allowing it to dynamically adapt to diverse translation tasks without the need for extensive retraining. Extensive experiments on diverse datasets, using open LLM models such as Starcoder, Llama3-70B Instruct, CodeLlama-34B Instruct, Granite-34B Code Instruct, and Mixtral-8x22B, and commercial LLM models such as GPT-3.5 turbo, and GPT-4o demonstrate the superiority of our approach over traditional zero-shot, particularly in translating between Fortran and C++.We also explored different numbers of shots (examples provided to the model during inference) — specifically 1, 2, and 3 shots — and various embedding models for RAG, including Nomic-Embed, Starencoder, and CodeBERT, to evaluate the robustness and effectiveness of our approach.

\end{abstract}


\section{Introduction}
The rapid evolution of programming languages and the need to maintain legacy codebases have created a substantial demand for automated code translation tools. Traditional approaches to code translation involve extensive manual effort and expertise in both the source and target languages. With the rise of large language models (LLMs), such as GPT-3~\cite{roumeliotis2023chatgpt}, Codex~\cite{chen2021evaluating}, and CodeBERT~\cite{feng2020codebert}, automated code translation has become increasingly feasible. These models leverage vast amounts of training data to generate code snippets in various languages, demonstrating impressive capabilities in general language understanding and generation tasks.

However, despite these advancements, the performance of language models in code translation tasks remains inconsistent, particularly when dealing with complex or less common programming constructs. A significant challenge lies in the model's ability to comprehend and generate code that adheres to the syntactic and semantic rules of both the source and target languages. Zero-shot and few-shot learning techniques (where "shots" refer to the number of examples provided to the model during inference) have shown promise~\cite{kojima2022large}, but they often lack the depth of contextual understanding required for high-fidelity translations.

Fine-tuning language models for specific code translation tasks is a common approach to improve performance~\cite{LLNL2023}. However, fine-tuning requires substantial computational resources and time, and it may not generalize well to new or unseen tasks. In contrast, Retrieval-Augmented Generation (RAG) offers a dynamic alternative by leveraging a repository of existing code translations to provide contextual examples during the translation process~\cite{lewis2020retrieval}. This method allows the model to adapt to various tasks without extensive retraining, making it a more flexible and efficient solution.

In this paper, we propose a RAG framework  that enhances Few-Shot Learning for code translation tasks. Our approach involves maintaining a repository of code translation examples and dynamically retrieving the most relevant examples based on the input code segment. By providing the model with multiple contextual examples, RAG facilitates a deeper understanding of the translation task, leading to more accurate and reliable code generation.  For this, an embedding model, which converts code snippets into numerical vectors that capture their semantic meaning, is used.

We conducted extensive experiments using various open language models, including Starcoder \cite{li2023starcoder}, Llama3-70B Instruct \cite{llama3modelcard}, CodeLlama-34B Instruct \cite{roziere2023code}, Granite-34B Code Instruct \cite{mishra2024granite}, Mistral 8x22B \cite{jiang2024mixtral} and Codestral \cite{codestral}  and commercial models such as GPT-3.5 \cite{roumeliotis2023chatgpt} and GPT-4o on publicly available Fortran-C++ translation  datasets. We evaluated the performance of our approach with different numbers of shots (1, 2, 3) and various embedding models, such as Nomic-Embed \cite{nussbaum2024nomic}, Starencoder \cite{li2023starcoder}, and CodeBERT \cite{feng2020codebert}.  For datasets that did not have direct translation pairs, we performed pairwise comparisons to assess the translation consistency across different LLMs.

The remainder of this paper is structured as follows: Section 2 reviews related work in code translation and few-shot learning. Section 3 outlines our RAG methodology, including the retrieval mechanism and integration with language models. Section 4 presents our experimental setup and results, followed by a discussion in Section 5. Finally, Section 6 concludes the paper and highlights future research directions.

\section{Related works}
The field of code translation, particularly from Fortran to C++, has seen various innovative approaches and methodologies aimed at improving translation accuracy and efficiency. One notable contribution is the development of a high-performance computing (HPC) code translation dataset \cite{LLNL2023}. This dataset pairs OpenMP Fortran with C++ code snippets, facilitating the training and evaluation of machine learning models for effective code translation. The dataset's quality was ensured through human-level evaluations by expert programmers, which significantly refined the translations by assessing correctness, readability, and semantic retention of the original Fortran code.

In addition to dataset creation, efforts have been made to re-engineer legacy Fortran code into maintainable C++ code. A study  explored the maintainability of translated Fortran code by evaluating various software quality metrics \cite{Tomlinson2004}. The research concluded that re-engineered Fortran to C++ code exhibited high maintainability standards, making it a viable solution for modernizing legacy systems while ensuring the translated code remains efficient and easy to maintain.

Automated tools have also played a crucial role in facilitating Fortran to C++ translation. The CFortranTranslator, an open-source tool, converts Fortran90/Fortran77 code to C++14 while maintaining the abstraction level of the original code \cite{CFortranTranslator}. This tool effectively handles mixed fixed-form and free-form Fortran code, providing a practical solution for developers needing to translate Fortran code to leverage modern C++ features and frameworks; however, the resulting code may not always be easy to read or maintain, posing long-term challenges.

The advent of LLMs has further revolutionized code translation. A study titled ``Lost in Translation" emphasized the importance of providing contextual information, such as stack traces and error messages, to improve LLM-based code translation accuracy \cite{LostInTranslation2023}. This iterative prompting approach demonstrated significant improvements in translation success rates by incorporating additional informative context into the prompts. The study also provided a comprehensive taxonomy of bugs introduced by LLMs during code translation, offering valuable insights into common error modalities and mitigation strategies.

Fine-tuning LLMs for domain-specific tasks has been another area of active research \cite{FineTuning2023}. Prompt-oriented fine-tuning using Low-Rank Adaptation (LoRA) \cite{lora_paper} has been proposed to adapt LLMs to specific programming languages and domains. This method leverages domain-specific vocabulary and demonstrates improved translation quality by efficiently fine-tuning pre-trained models for specific tasks \cite{LLNL2023}. In the paper, the authors demonstrated that finetuning the LLM improved the translation performance by 9 folds.

\section{Methods}
\begin{figure}[ht!]
    \centering
    \includegraphics[width=0.5\textwidth]{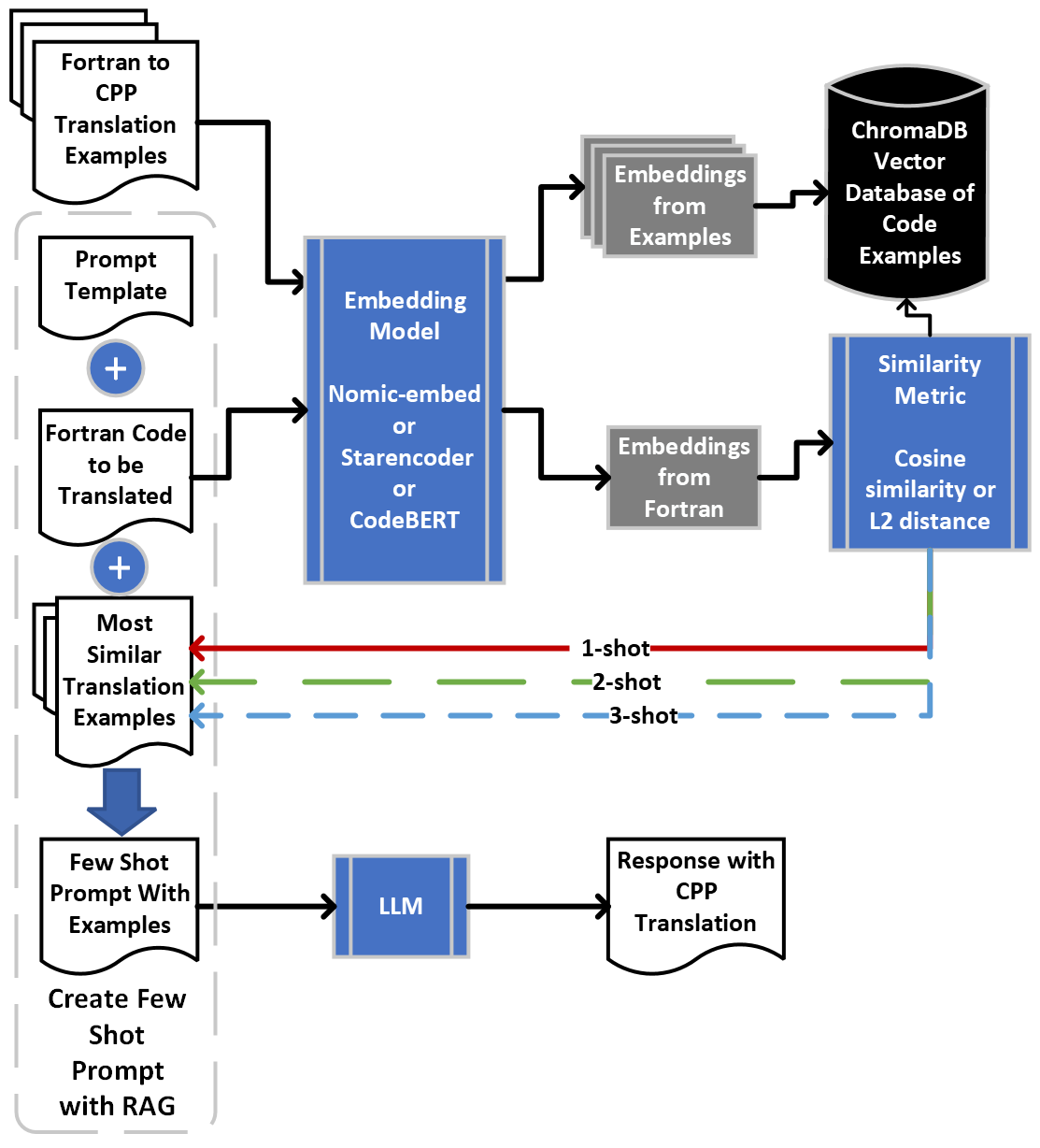}
    \caption{Pipeline for creating a few shot prompt through RAG for code translation.}
    \label{fig:overview}
\end{figure}
Our study presents a RAG based pipeline designed to enhance the accuracy and contextual understanding of automated code translations from Fortran to C++, as illustrated in Figure \ref{fig:overview}. This method integrates LLMs with retrieval mechanisms, enabling the generation of high-quality translations through dynamically provided contextual examples. The pipeline is highly adaptive, allowing users to plug in different LLMs, embedding models, datasets, and adjust the number of shots for evaluating translation performance. Models can be seamlessly loaded directly from Hugging Face \cite{wolf2020huggingfacestransformersstateoftheartnatural}, specific directories tailored to each model type, or via API calls as specified by user input. For retrieval with RAG, either $l_2$ distance or $cosine$ similarity metric can be used.

\subsection*{Dataset Preparation}

We leveraged three datasets to evaluate our models:

\subsubsection{Numerical Recipes Dataset:} This dataset comprises pairs of Fortran and C++ code snippets, ensuring a robust set of examples \cite{press1988numerical}. Each code pair is meticulously curated to maintain high standards of quality and relevance. To ensure quality, we standardized code style, removed comments, handled whitespace and special characters, and mapped Fortran subroutines to their C++ equivalents. One downside of this dataset is that it relies on a specific library of functions, which may limit its general applicability. This dataset comprises 298 Fortran-C++ pairs.

\subsubsection{HPC Fortran2CPP Dataset:} This dataset was derived from \cite{LLNL2023}.The dataset comprises comprehensive Fortran to C++ translation pairs and  was meticulously curated from the NAS Parallel Benchmarks (NPB), Polyhedral Benchmark (PolyBench), and DataRaceBench (DRB) repositories. The NPB dataset evaluates supercomputer performance with computational fluid dynamics benchmarks, PolyBench provides programs for polyhedral compilation research, and DRB includes OpenMP programs for data race detection tool evaluation. The authors performed standardized code style as we did for Numerical Recipe dataset and additional calibration was done using similarity tests ensured semantic fidelity, while expert programmers conducted human-level evaluations to assess correctness, readability, and semantic retention. This dataset comprises 315 Fortran-C++ pairs.
\subsubsection{Stack-V2 Dataset:} The Stack V2 dataset is a comprehensive collection of code samples sourced from a wide range of repositories on GitHub, focusing on high-performance computing and various computational problems \cite{lozhkov2024starcoder}. This dataset includes approximately half a million Fortran code snippets, providing a diverse set of examples for robust evaluation. For our study, we sampled 500 Fortran examples from this dataset by selecting files with lengths between 1000 and 10,000 bytes from unique repositories, prioritizing those with the highest combined star and fork event counts to ensure high-quality and diverse samples. Since Stack-V2 doesn't have Fortran-C++ pairs, we extracted the files containing metadata, codes, and comments. We then leveraged Llama3-70B Instruct model to extract the executable Fortran code, discarding other metadata. This cleaned dataset was subsequently used to compare translation performance across various LLMs in a zero-shot setting.

\subsection*{Embedding Generation and Example Retrieval}
\begin{figure}[ht!]
    \centering
    \begin{subfigure}[b]{0.49\linewidth}
        \includegraphics[width=\linewidth]{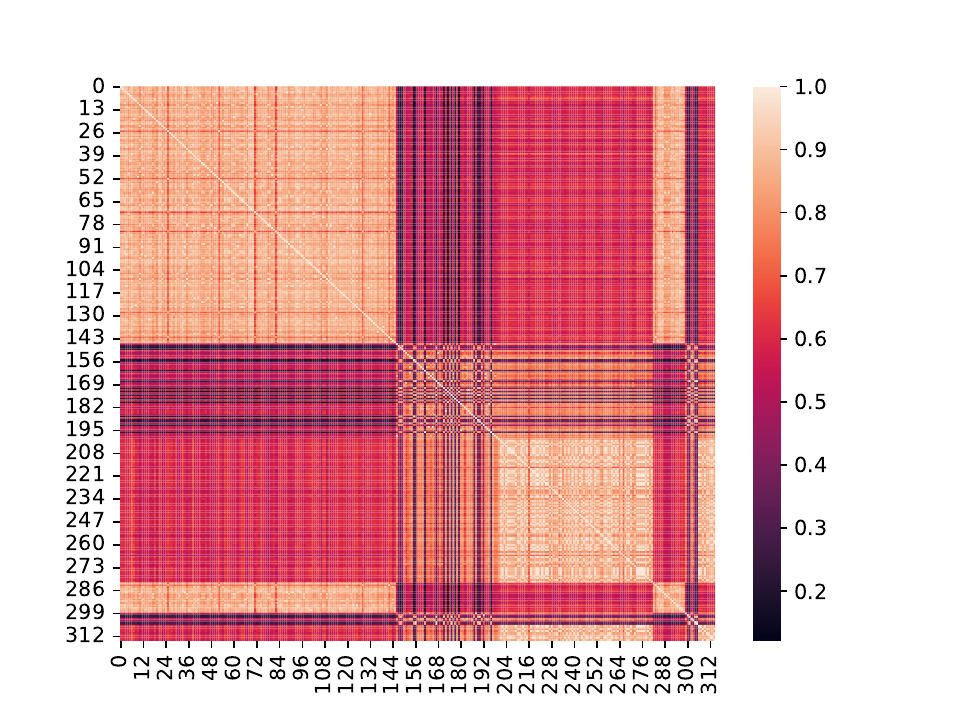}
        \caption{}
        \label{fig:sub1}
    \end{subfigure}
    \begin{subfigure}[b]{0.49\linewidth}
        \includegraphics[width=\linewidth]{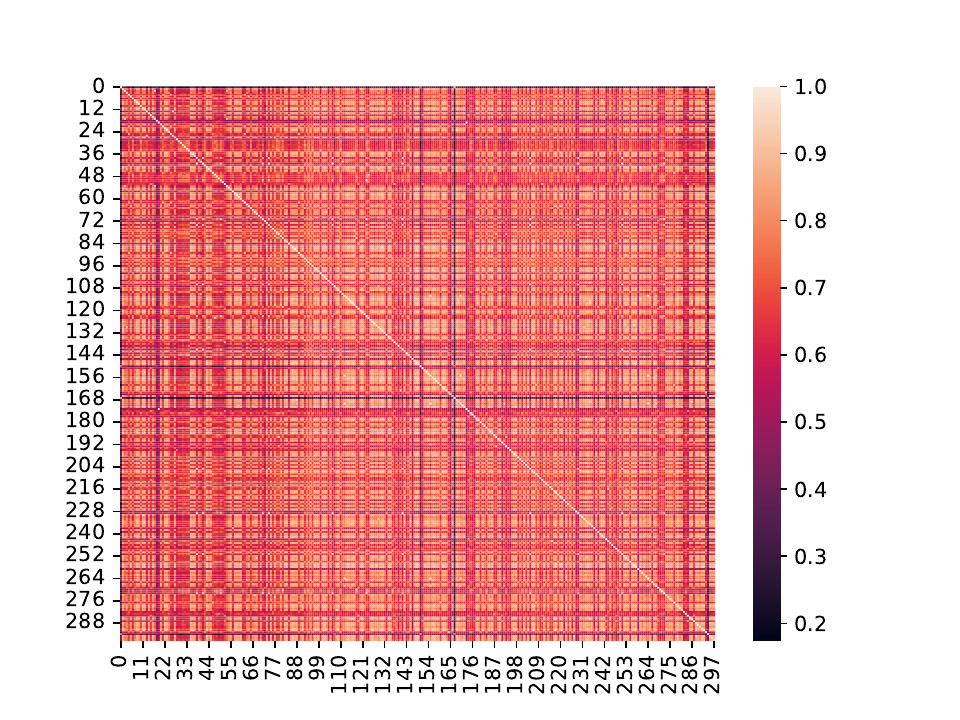}
        \caption{}
        \label{fig:sub2}
    \end{subfigure}
    \caption{Similarity of dataset embeddings for a)HPC Fortran2CPP dataset and b) Numerical Receipe dataset based on Nomic Embed model}
    \label{fig:test}
\end{figure}
A crucial component of our RAG pipeline is the generation of embeddings for each Fortran code snippet. We employ various embedding models such as Nomic-Embed \cite{nussbaum2024nomic}, Starencoder \cite{li2023starcoder}, and CodeBERT \cite{feng2020codebert} to produce these embeddings, which are essential for assessing translation performance. These embeddings capture the semantic essence of the code snippets, enabling efficient retrieval of the most contextually relevant examples from the dataset. For efficient vector storage and retrieval, we leverage ChromaDB, which stores embeddings along with the corresponding source and target codes. We evaluated the retrieval performances with $l_2$ and $cosine$ similarity metrics.

Once the embeddings are generated, the system dynamically retrieves the most similar examples based on cosine similarity or Euclidean distance. This process ensures that the model receives the most relevant examples tailored to each specific translation task, thereby enhancing the accuracy and efficiency of the code translation. The retreived translation pairs are then augmented with the original query and directed to LLM for translation.
Figure \ref{fig:test} displays the cosine similarity between the Fortran source code embeddings corresponding to the Numerical Recipes and HPC Fortran2CPP datasets. The similarity matrices provide valuable insights into the structural and functional relationships within and between the datasets.

In the HPC Fortran2CPP dataset (Figure \ref{fig:sub1}), we observe distinct clusters of high similarity, indicating that certain groups of Fortran code snippets share significant semantic and syntactic features. These clusters likely correspond to common computational patterns or routines that are frequently reused within the dataset. The clear block-diagonal structure suggests that the HPC Fortran2CPP dataset contains several distinct modules or subcomponents that exhibit high internal coherence. In contrast, the Numerical Recipes dataset (Figure \ref{fig:sub2}) presents a more uniformly distributed similarity matrix with fewer pronounced clusters. This distribution implies a more diverse set of code snippets with varying functionalities and less repetition of specific patterns. The relatively consistent similarity across the matrix suggests that while individual snippets may not be as closely related, there is a broad underlying similarity in the type of computational problems addressed by the Numerical Recipes dataset.

Comparing the two datasets, we can infer that the HPC Fortran2CPP dataset might be more modular and specialized, with specific sections of code dedicated to particular tasks. In contrast, the Numerical Recipes dataset appears to be more general-purpose, covering a wide range of computational routines without strong modular separation. These insights are crucial for understanding the nature of the datasets and for accessing the RAG performance for code translation, as they highlight the varying degrees of internal consistency and the potential challenges in translating diverse code structures.

\subsection*{Few-Shot Learning with Retrieval-Augmented Generation}

In-context learning  for source code translation leverages the power of few-shot examples to improve the performance of language models in generating accurate translations. In a typical zero-shot setting, a language model \( M \) generates a target translation \( T \) for a given query \( Q \) (source code) directly, which can be mathematically represented as \( T = M(Q) \). However, the performance of \( M \) can be significantly enhanced by conditioning it on a set of \( k \) example pairs of source and target code, \(\{(S_i, T_i)\}_{i=1}^{k}\), before generating the translation for \( Q \). This few-shot learning process can be expressed as:

\[
T = M(\{(S_i, T_i)\}_{i=1}^{k}, Q).
\]

In a RAG setup, this process is further optimized by incorporating a retrieval mechanism \( R \) that selects the most relevant \( k \) example pairs from a large corpus \( \mathcal{C} \) based on the query \( Q \). The retrieval step can be mathematically formulated as:

\[
\{(S_i, T_i)\}_{i=1}^{k} = R(Q, \mathcal{C}).
\]

Subsequently, the model \( M \) generates the translation \( T \) using the retrieved example pairs and the query, expressed as:

\[
T = M(R(Q, \mathcal{C}), Q).
\]

The translation process begins with a zero-shot approach, where the model translates the Fortran code to C++ without any additional context. We then generate embeddings for the Fortran code snippets using models like Nomic-Embed, Starencoder, and CodeBERT. These embeddings capture the semantic essence of the code, facilitating efficient retrieval of the top-k most relevant examples from ChromaDB, our vector storage and retrieval system.


\begin{figure}
\begin{lstlisting}[numbers=none]
System: You are adept at translating Fortran code into CPP with high accuracy, ensuring that all syntax, semantics, and specific language features are correctly and efficiently converted
User: Translate the following code from Fortran to CPP:
{Fortran code snippet}
\end{lstlisting}
\caption{Zero-Shot Translation Prompt Template}
\label{fig:zero-shot-translation-prompt-template}
\end{figure}
\begin{figure}
\begin{lstlisting}[numbers=none]
System: You are adept at translating Fortran code into CPP with high accuracy, ensuring that all syntax, semantics, and specific language features are correctly and efficiently converted
User: Here's an example of a code translated from Fortran to CPP:
Here's the Fortran code:
{Similar Fortran code example}
Here's the CPP translation:
{Corresponding CPP translation}
[Repeat for k examples]
Now translate the following code from Fortran to CPP:
{Fortran code snippet}
\end{lstlisting}
\caption{Few Shot Translation Prompt Template}
\label{fig:few-shot-translation-prompt-template}
\end{figure}

For Zero-shot comparisons, we apply the zero-shot translation prompt template (Figure \ref{fig:zero-shot-translation-prompt-template}) with the Fortran code to be translated. For few-shot prompts, we use the prompt template shown in Figure \ref{fig:few-shot-translation-prompt-template} with the rest of the pipeline. The model processes the input Fortran code along with the retrieved examples. The tokenized input is fed into the model, which generates the corresponding C++ code. This approach ensures that the model receives the most relevant examples tailored to each specific translation task, thereby enhancing translation accuracy and efficiency.

\begin{figure*}[ht!]
    \centering
    \begin{subfigure}[b]{0.32\linewidth}
            \includegraphics[width=\linewidth]{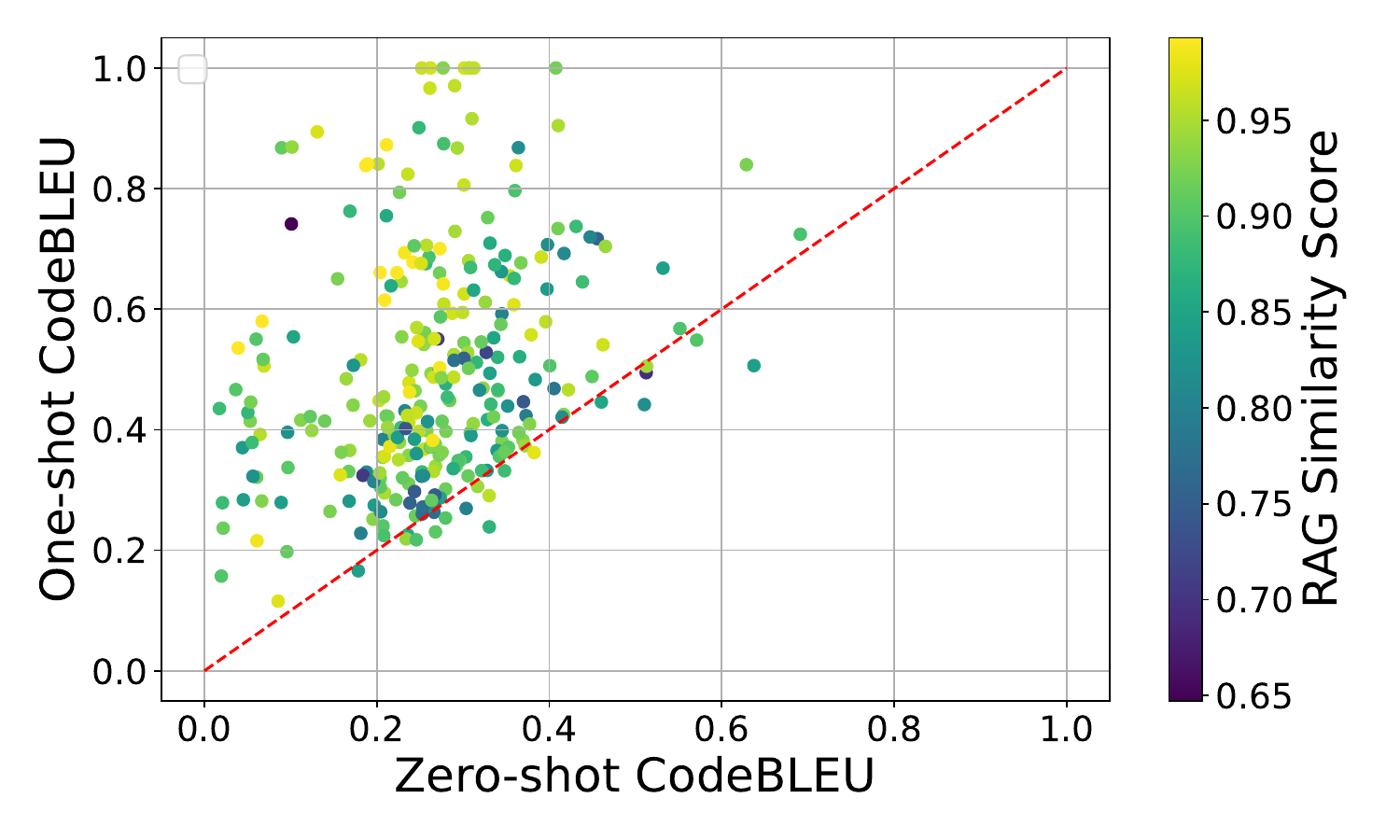}

        \caption{}
        \label{fig:sub11}
    \end{subfigure}
    \hfill 
    \begin{subfigure}[b]{0.32\linewidth}
        \includegraphics[width=\linewidth]{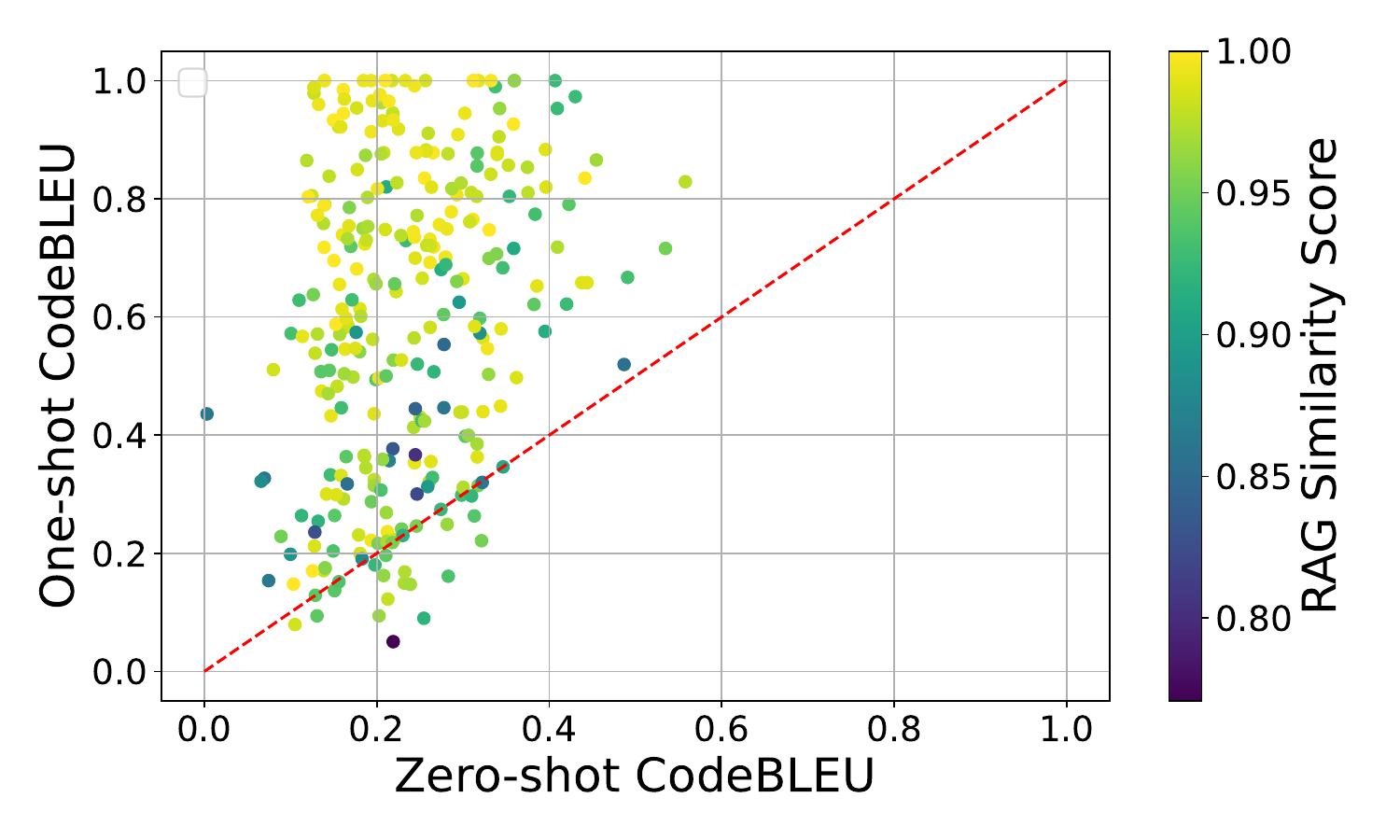}
        \caption{}
        \label{fig:sub12}
    \end{subfigure}
    \hfill 
    \begin{subfigure}[b]{0.32\linewidth}
    \includegraphics[width=\linewidth]{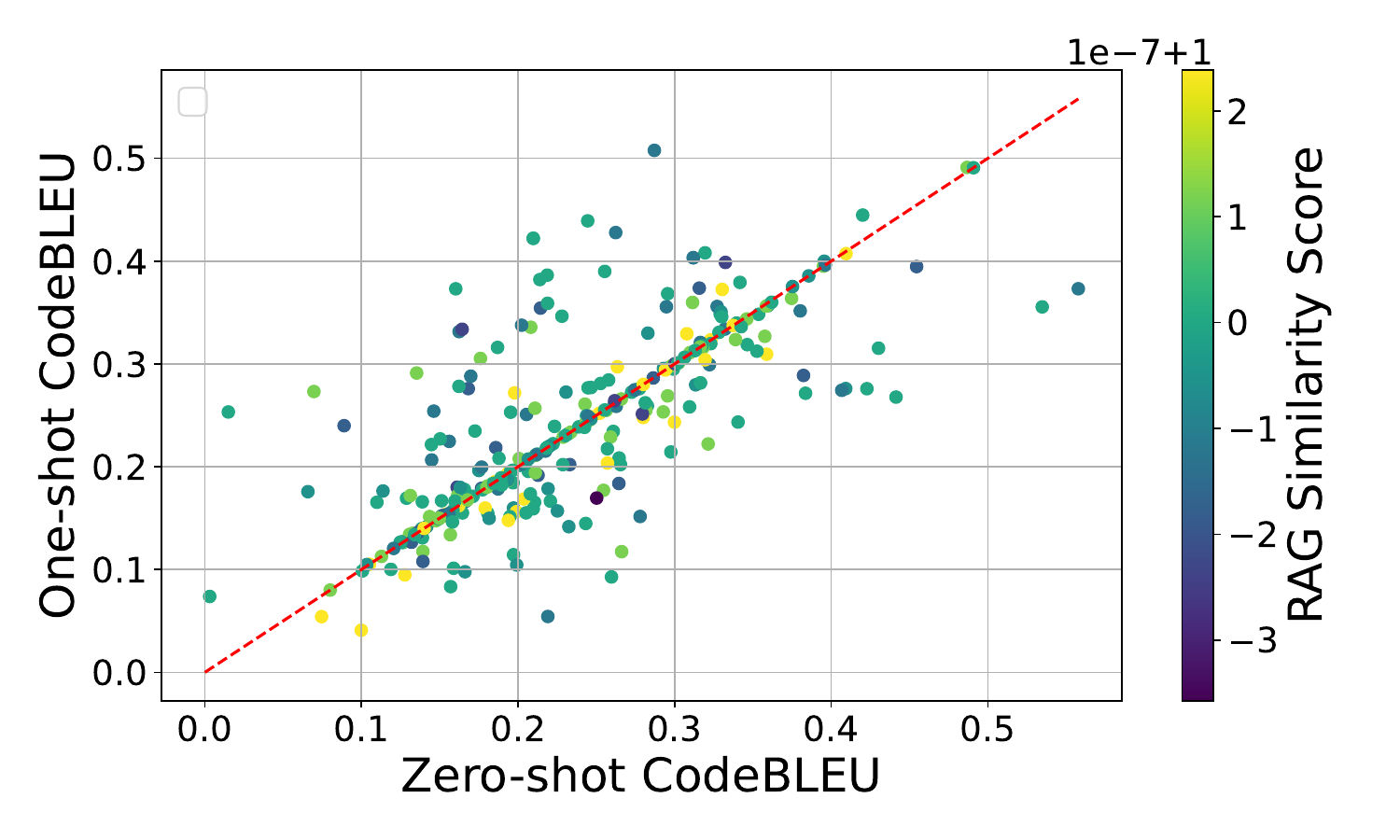}
        \caption{}
        \label{fig:sub13}
    \end{subfigure}
\caption{Performance Comparison of One-shot vs. Zero-shot in the RAG Pipeline Using the Nomic-embed Embedding Model across Various Models and Datasets: (a) Granite-34B Code Instruct on the Numerical Recipes Dataset, (b) Granite-34B Code Instruct on the HPC Fortran2CPP Dataset, and (c) Granite-34B Code Instruct on the HPC Fortran2CPP Dataset with bad RAG setup (utilize largest distance metric as retreival) . The color of each data point represents the similarity of the retrieved one-shot example pair to the query Fortran code, with the legend indicating the intensity range of the similarity metric. Generally, a higher similarity score correlates with a higher CodeBLEU score.}

    \label{fig:test2}
\end{figure*}
\subsection*{Evaluation and Experimental Setup}

The generated translations are evaluated using the CodeBLEU metric~\cite{ren2020codebleu}, which assesses the quality of code translations by considering both syntactic and semantic correctness. CodeBLEU extends the traditional BLEU metric by incorporating four key components: \textit{N-gram Match Score}, which measures the precision of n-grams in the translated code compared to the reference code, ensuring the retention of original token sequences; \textit{Weighted N-gram Match Score}, which enhances the n-gram match score by weighting different n-grams based on their importance, thus focusing on critical code patterns and structures; \textit{Syntax Match Score}, which evaluates the syntactic correctness of the translated code, ensuring adherence to the grammatical rules of the programming language; and \textit{Dataflow Match Score}, which assesses the semantic correctness by analyzing the data flow within the program, ensuring the preservation of logical flow and functional equivalence. 

Our experiments involved extensive testing with various combinations of models, shot numbers, RAG retrieval metrics, and embedding models. We utilized Nomic-Embed, Starencoder, and CodeBERT as embedding models. For code translation, we leveraged open LLMs such as Starcoder, Llama3-70B Instruct, Code LLaMA-34B, Granite-34B Code Instruct, and Mistral 8x22B, as well as commercial LLM models such as GPT-3.5 turbo and GPT-4o/GPT-4 turbo. We evaluated the performance of the models for zero, one, two, and three shots. Each experiment aimed to measure the improvements in translation quality facilitated by the RAG approach using CodeBLEU as the primary evaluation metric.

\section{Results and Discussions}

\begin{figure*}[!ht]
\centering

\begin{subfigure}[b]{.49\textwidth}
  \centering
  \includegraphics[width=\linewidth]{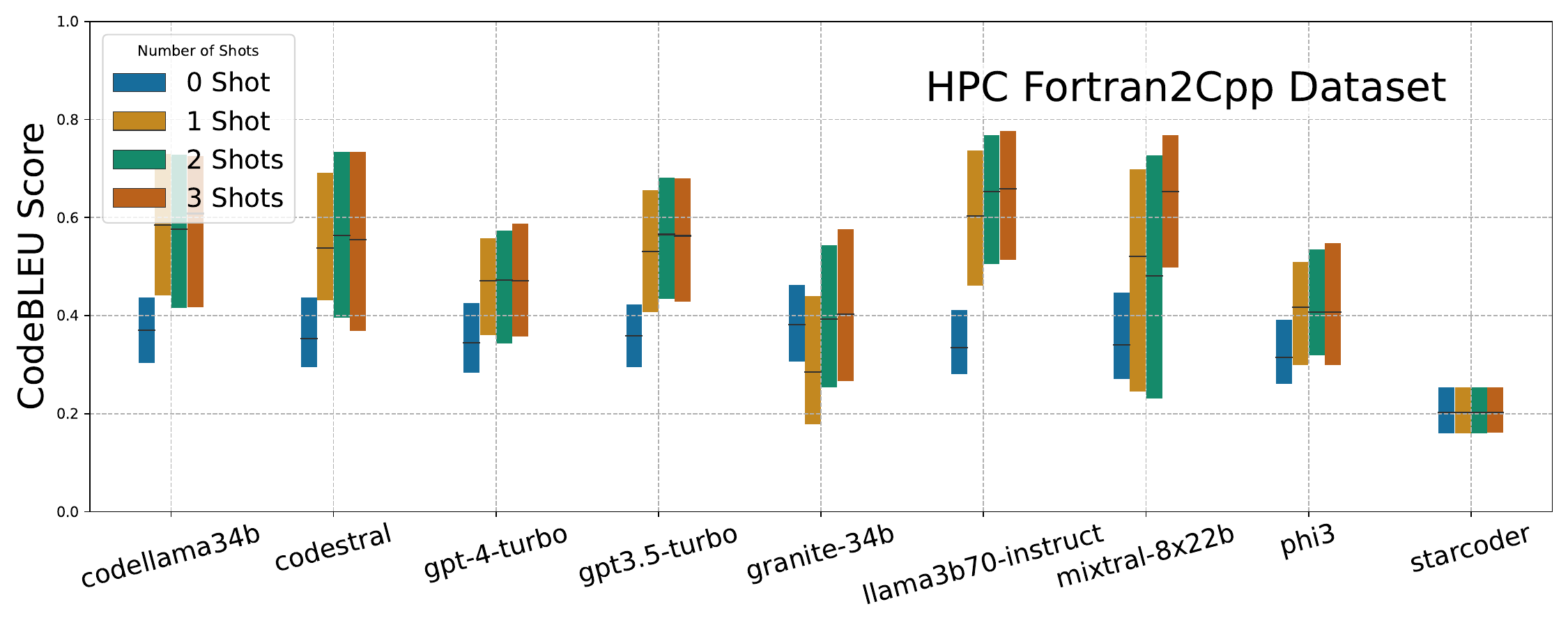}
  \caption{HPC Fortran2CPP dataset with cosine similarity}
  \label{fig:cos_llnl}
\end{subfigure}
\hfill
\begin{subfigure}[b]{.49\textwidth}
  \centering
  \includegraphics[width=\linewidth]{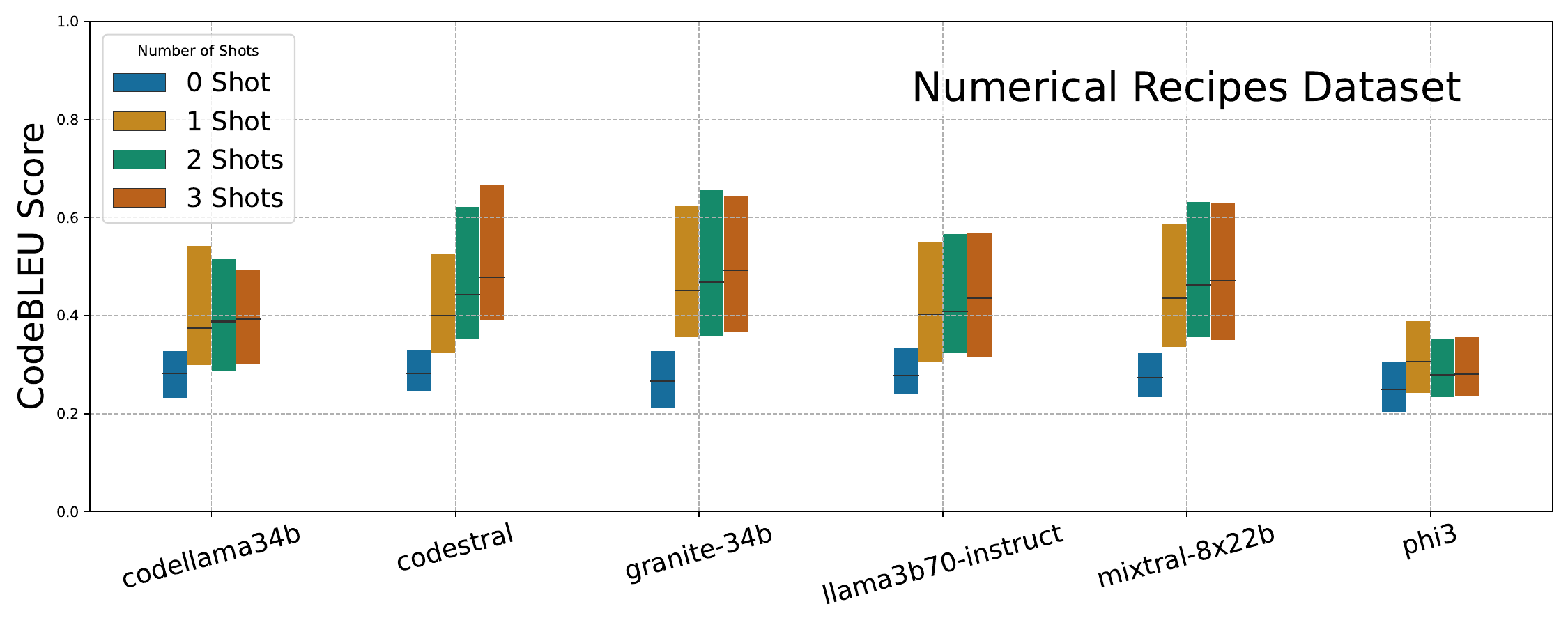}
  \caption{Numerical Recipes dataset with cosine similarity}
  \label{fig:cos_ncp}
\end{subfigure}

\vspace{0.5cm} 

\begin{subfigure}[b]{.49\textwidth}
  \centering
  \includegraphics[width=\linewidth]{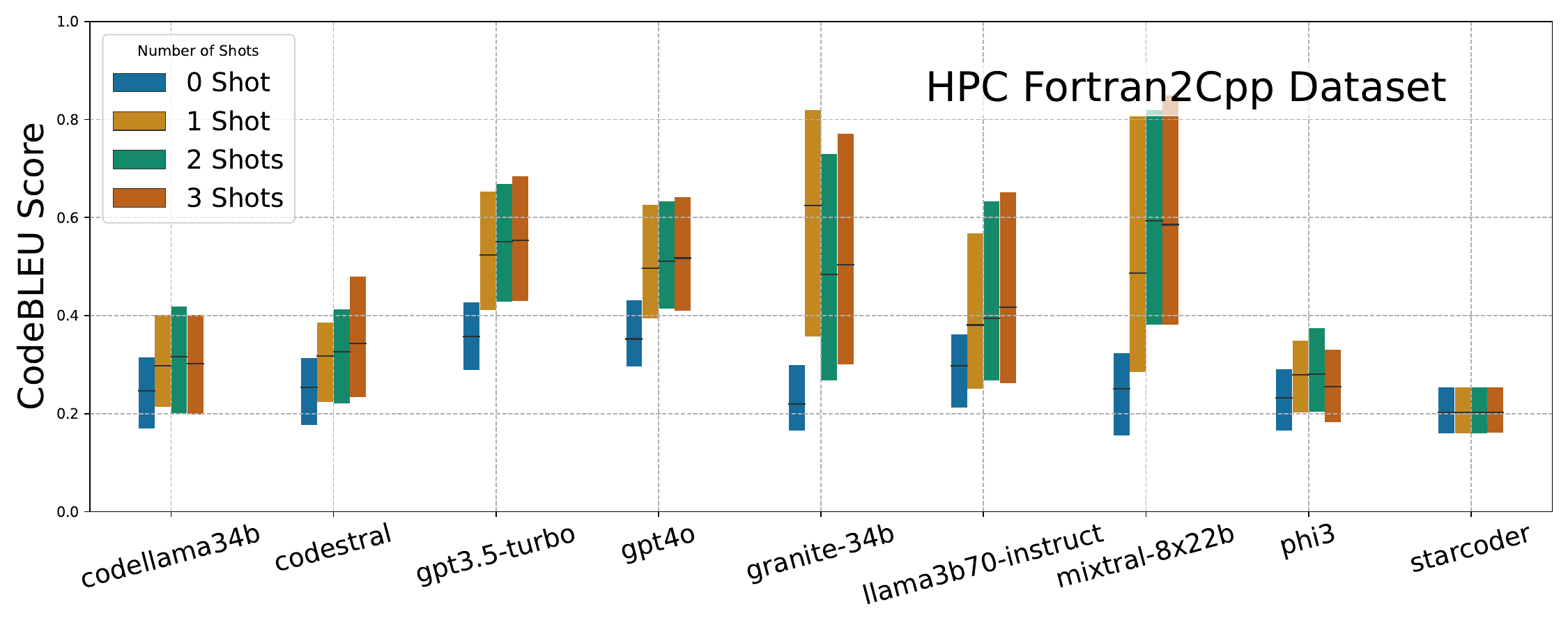}
  \caption{HPC Fortran2CPP dataset with $l_2$ distance}
  \label{fig:l2_llnl}
\end{subfigure}
\hfill
\begin{subfigure}[b]{.49\textwidth}
  \centering
  \includegraphics[width=\linewidth]{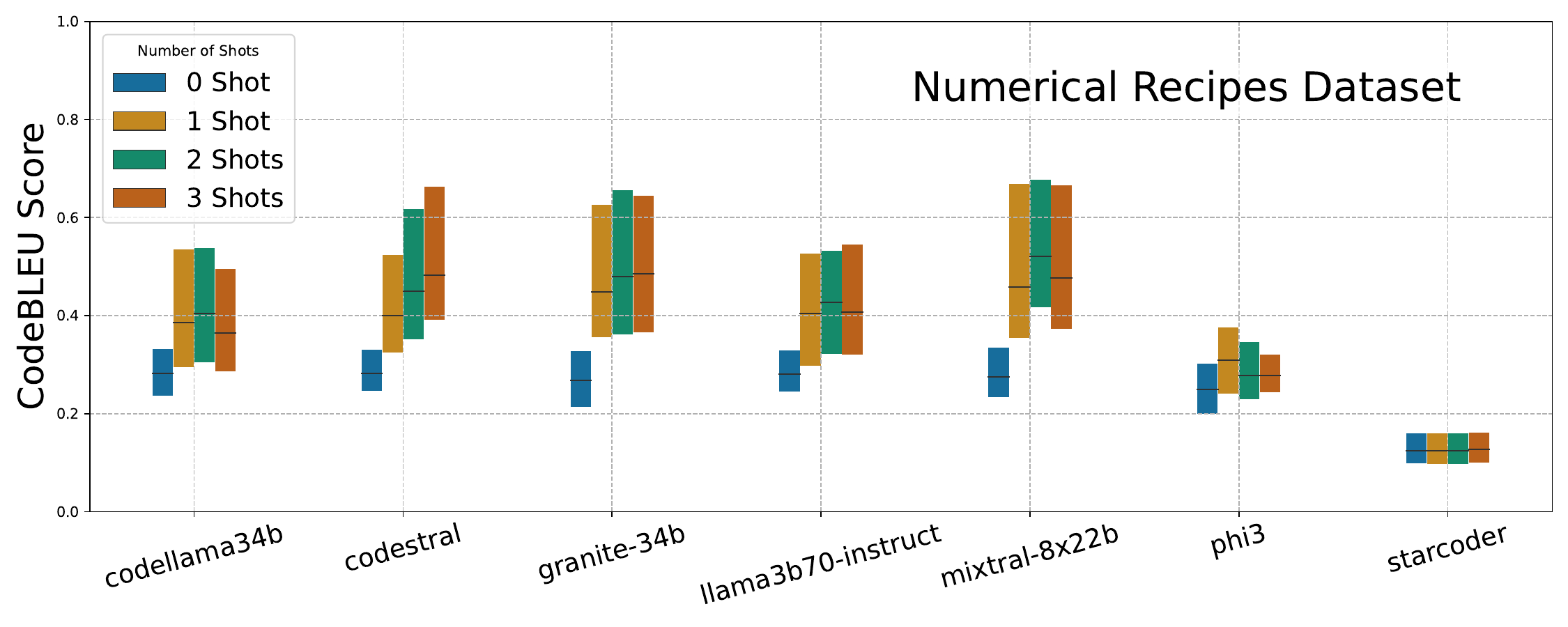}
  \caption{Numerical Recipes dataset with $l_2$ distance}
  \label{fig:l2_ncp}
\end{subfigure}

\caption{Overview of CodeBLEU metrics for code translation with different LLMs using cosine similarity and $l_2$ distance for RAG retrieval on (a) and (c) HPC Fortran2CPP dataset and (b) and (d) Numerical Recipes dataset.}
\label{fig:codebleu_overview}
\end{figure*}

\begin{table*}[h]
\caption{Mean with standard deviation scores for CodeBLEU metrics of Zero Shot prompts for different LLMs, ordered by CodeBLEU performance}
\label{tab:mean_sem_codebleu_nomic_embed_cosine_0}

\centering
\begin{tabular}{lllllll}
\toprule
\textbf{Dataset} & \textbf{Model} & \textbf{CodeBLEU} & \textbf{Ngram} & \textbf{Weighted Ngram} & \textbf{Syntax Tree} & \textbf{Dataflow} \\ \toprule 

HPC Fortran2CPP & GPT-4o & $0.371 \pm 0.002$ & $0.188 \pm 0.001$ & $0.290 \pm 0.001$ & $0.495 \pm 0.003$ & $0.504 \pm 0.005$ \\ 
 & GPT-3.5 Turbo & $0.367 \pm 0.001$ & $0.199 \pm 0.003$ & $0.301 \pm 0.004$ & $0.483 \pm 0.002$ & $0.474 \pm 0.004$ \\ 
 & Llama3-70B Instruct & $0.309 \pm 0.001$ & $0.135 \pm 0.002$ & $0.225 \pm 0.002$ & $0.325 \pm 0.005$ & $0.512 \pm 0.003$ \\ 
 & Codestral-22B & $0.245 \pm 0.000$ & $0.090 \pm 0.001$ & $0.155 \pm 0.001$ & $0.158 \pm 0.002$ & $0.520 \pm 0.002$ \\ 
 & CodeLlama-34B Instruct & $0.243 \pm 0.000$ & $0.090 \pm 0.001$ & $0.166 \pm 0.001$ & $0.169 \pm 0.002$ & $0.490 \pm 0.001$ \\ 
 & Mixtral-8x22B (176B) & $0.241 \pm 0.001$ & $0.059 \pm 0.001$ & $0.104 \pm 0.001$ & $0.140 \pm 0.001$ & $0.331 \pm 0.001$ \\ 
 & Granite-34B Code Instruct & $0.237 \pm 0.001$ & $0.084 \pm 0.001$ & $0.145 \pm 0.001$ & $0.171 \pm 0.000$ & $0.500 \pm 0.002$ \\ 
 & Phi-3 3.8B & $0.228 \pm 0.000$ & $0.063 \pm 0.004$ & $0.119 \pm 0.006$ & $0.165 \pm 0.001$ & $0.501 \pm 0.002$ \\ 
 & StarCoder 15.5B & $0.206 \pm 0.000$ & $0.009 \pm 0.000$ & $0.014 \pm 0.000$ & $0.127 \pm 0.000$ & $0.636 \pm 0.001$ \\ 

\hline 

Numerical Recipes & Codestral-22B & $0.288 \pm 0.000$ & $0.042 \pm 0.000$ & $0.101 \pm 0.000$ & $0.512 \pm 0.000$ & $0.497 \pm 0.000$ \\ 
 & Llama3-70B Instruct & $0.283 \pm 0.001$ & $0.036 \pm 0.000$ & $0.103 \pm 0.001$ & $0.508 \pm 0.003$ & $0.484 \pm 0.002$ \\ 
 & CodeLlama-34B Instruct & $0.281 \pm 0.002$ & $0.035 \pm 0.001$ & $0.108 \pm 0.003$ & $0.482 \pm 0.007$ & $0.493 \pm 0.003$ \\ 
 & Granite-34B Code Instruct & $0.272 \pm 0.002$ & $0.063 \pm 0.001$ & $0.109 \pm 0.001$ & $0.462 \pm 0.003$ & $0.443 \pm 0.003$ \\ 
 & Mixtral-8x22B (176B) & $0.264 \pm 0.021$ & $0.077 \pm 0.015$ & $0.140 \pm 0.013$ & $0.340 \pm 0.070$ & $0.385 \pm 0.052$ \\ 
 & Phi-3 3.8B & $0.249 \pm 0.002$ & $0.036 \pm 0.001$ & $0.085 \pm 0.002$ & $0.432 \pm 0.002$ & $0.438 \pm 0.004$ \\ 
 & StarCoder 15.5B & $0.133 \pm 0.001$ & $0.007 \pm 0.000$ & $0.010 \pm 0.000$ & $0.269 \pm 0.000$ & $0.244 \pm 0.002$ \\ 

\bottomrule
\end{tabular}
\end{table*}

\begin{table*}[!ht]
\caption{Delta in Mean CodeBLEU scores between Zero- and Few-Shot prompts using the Nomic-embed Embedding Model with $l_2$ Distance for HPC Fortran2CPP dataset. Our RAG method for choosing examples improves CodeBLEU scores for most of the LLMs listed. }
\label{tab:mean_sem_codebleu_nomic_embed_l2}

\centering
\begin{tabular}{lllllll}
\toprule
& & \multicolumn{4}{c}{}{\textbf{$\Delta$ in CodeBLEU scores }} \\ 
\cmidrule{4-6}
\textbf{Dataset} & \textbf{Model} & \textbf{Zero-shot} & \textbf{1-shot} & \textbf{2-shot} & \textbf{3-shot} \\ \toprule 

HPC Fortran2CPP & CodeLlama-34B Instruct & 0.243 & +0.078 & +0.084 & +0.069 \\ 
 & Codestral-22B & 0.245 & +0.074 & +0.105 & +0.158 \\ 
 & GPT-3.5 Turbo & 0.367 & +0.157 & +0.176 & +0.188 \\ 
 & GPT-4o & 0.371 & +0.132 & +0.153 & +0.153 \\ 
 & Granite-34B Code Instruct & 0.237 & \textbf{+0.363} & +0.278 & +0.302 \\ 
 & Llama3-70B Instruct & 0.309 & +0.117 & +0.137 & +0.151 \\ 
 & Mixtral-8x22B (176B) & 0.241 & +0.288 & \textbf{+0.355} & \textbf{+0.367} \\ 
 & Phi-3 3.8B & 0.228 & +0.058 & +0.070 & +0.050 \\ 
 & StarCoder 15.5B & 0.206 & 0.000 & 0.000 & 0.000 \\ 

\hline 

Numerical Recipes & CodeLlama-34B Instruct & 0.281 & +0.145 & +0.157 & +0.147 \\ 
 & Codestral-22B & 0.288 & +0.148 & +0.207 & +0.250 \\ 
 & Granite-34B & 0.272 & \textbf{+0.229} & +0.240 & +0.254 \\ 
 & Llama3-70B Instruct & 0.283 & +0.149 & +0.163 & +0.168 \\ 
 & Mixtral-8x22B (176B) & 0.264 & +0.227 & \textbf{+0.264} & \textbf{+0.291} \\ 
 & Phi-3 3.8B & 0.249 & +0.068 & +0.051 & +0.044 \\ 
 & StarCoder 15.5B & 0.133 & 0.000 & 0.000 & 0.000 \\ 

\bottomrule
\end{tabular}
\end{table*}


\subsubsection{Performance Across Models and Embeddings}
The results highlight significant variability in performance across different models and embedding strategies. The breakdown of Codebleu metric for different models in Zero-shot setting is shown in Table \ref{tab:mean_sem_codebleu_nomic_embed_cosine_0}. The table provides key insights into the performance of different models on zero-shot tasks using the Nomic-embed embedding model with $l_2$ distance. Notably, GPT-4 Turbo and GPT-3.5 Turbo achieved the highest zero-shot CodeBLEU scores of 0.371 and 0.367, respectively, indicating their strong initial performance without additional context. These models also scored highly on Syntax Tree and Dataflow metrics, which measure syntactic correctness and dataflow consistency, respectively, highlighting their ability to generate structurally sound and semantically accurate code. In contrast, Granite-34B and Llama3-70B Instruct, while having slightly lower CodeBLEU scores (0.237 and 0.309, respectively), showed balanced performance across all metrics, signifying their robust handling of various translation aspects. Metrics like Ngram and Weighted Ngram, which assess exact token matches and weighted token matches, respectively, were particularly low for StarCoder, reflecting its struggles with precise code generation in zero-shot tasks. Mixtral-8x22B exhibited high variance in the Numerical Recipes dataset, especially in CodeBLEU and Ngram scores, suggesting inconsistency in its translation quality. 

However in the Few-shot task, the Granite-34B Code Instruct, Llama3-70B Instruct and Mixtral-8x22B model consistently outperformed others across all embedding types and learning configurations, achieving the highest CodeBLEU scores. For instance, under the nomic-embed model embedding, Granite-34B Code Instruct achieved a zero-shot CodeBLEU of $0.237 $ and improved to $0.6$ in the one-shot setting for HPC Fortran2CPP dataset with $l_2$ norm as shown in Table \ref{tab:mean_sem_codebleu_nomic_embed_l2}. This demonstrates the model's strong capability in understanding and translating Fortran code into C++ with minimal context.

While we conducted experiments using various embedding models, our findings indicate that Nomic-embed and Starencoder exhibited equivalent performance in few-shot settings. However, CodeBERT consistently underperformed compared to the other two models. For instance, CodeLlama-34B Instruct achieved a zero-shot CodeBLEU of 0.243 with Nomic-embed, which improved to 0.321 in the two-shot configuration. In contrast, CodeBERT's performance did not show comparable improvement. This discrepancy can be attributed to the maximum token limit of each embedding model; CodeBERT has a token limit of 512, whereas both Starencoder and Nomic-embed support up to 8192 tokens. Given the longer length of Fortran codes to be translated, the limited token capacity of CodeBERT likely hindered its performance. To ensure clarity and focus, we present results using Nomic-embed throughout the text, as they are consistent with those obtained using Starencoder.

\subsubsection{Impact of Few-Shot Learning}

Few-shot learning consistently enhanced translation performance across all models and embeddings. The progressive improvement from one-shot to three-shot learning configurations was evident, with models like Granite-34B Code Instruct, Llama3-70B Instruct, and Mixtral-8x22B showing substantial gains as shown in Table~\ref{tab:mean_sem_codebleu_nomic_embed_l2}. For instance, Granite-34B Code Instruct with nomic\_embed improved from a zero-shot CodeBLEU of $0.24 \pm 0.09$ to $0.60 \pm 0.27$ in the one-shot setting, further improving to $0.54 ± 0.27$ and $0.54 \pm 0.21$ in the two-shot and three-shot settings, respectively. Thus, the RAG setup played a significant role in enhancing code translation quality. By leveraging a retrieval mechanism to provide relevant context from similar code snippets or documentation, RAG setups help models to better understand the structure and semantics of the source code. This additional context is especially beneficial in few-shot learning scenarios, where providing relevant examples can significantly improve the translation quality.

Figure \ref{fig:test2} illustrates the performance comparison of one-shot versus zero-shot scenarios in the RAG pipeline using the Nomic-embed embedding model across various models and datasets. Specifically, the comparisons are made for Figure \ref{fig:sub11} Granite-34B Code Instruct on the Numerical Recipes Dataset, Figure \ref{fig:sub12} Granite-34B Code Instruct on the HPC Fortran2CPP Dataset, and Figure \ref{fig:sub13} Granite-34B Code Instruct on the HPC Fortran2CPP Dataset with a poorly configured RAG setup (utilizing the largest distance metric as retrieval). The color of each data point represents the similarity of the retrieved one-shot example pair to the query Fortran code, with the legend indicating the intensity range of the similarity metric. Generally, the plots demonstrate that a higher similarity score between the retrieved example and the query correlates with a higher CodeBLEU score. When the utilized retrieval metric is altered, this brings a significant change in performances. To further validate this, we have leveraged $l_2$ distance and cosine similarity as retrieval metric. Figure~\ref{fig:l2_llnl} and ~\ref{fig:l2_ncp} showcases model performances leveraging $l_2$ metric whereas Figure~\ref{fig:cos_llnl} and ~\ref{fig:cos_ncp} showcases performance with cosine similarity. This relationship underscores the effectiveness of leveraging similar examples in enhancing the performance of the RAG pipeline, particularly in one-shot scenarios. 
\begin{figure}[!ht]
    \centering
    \includegraphics[width=\linewidth]{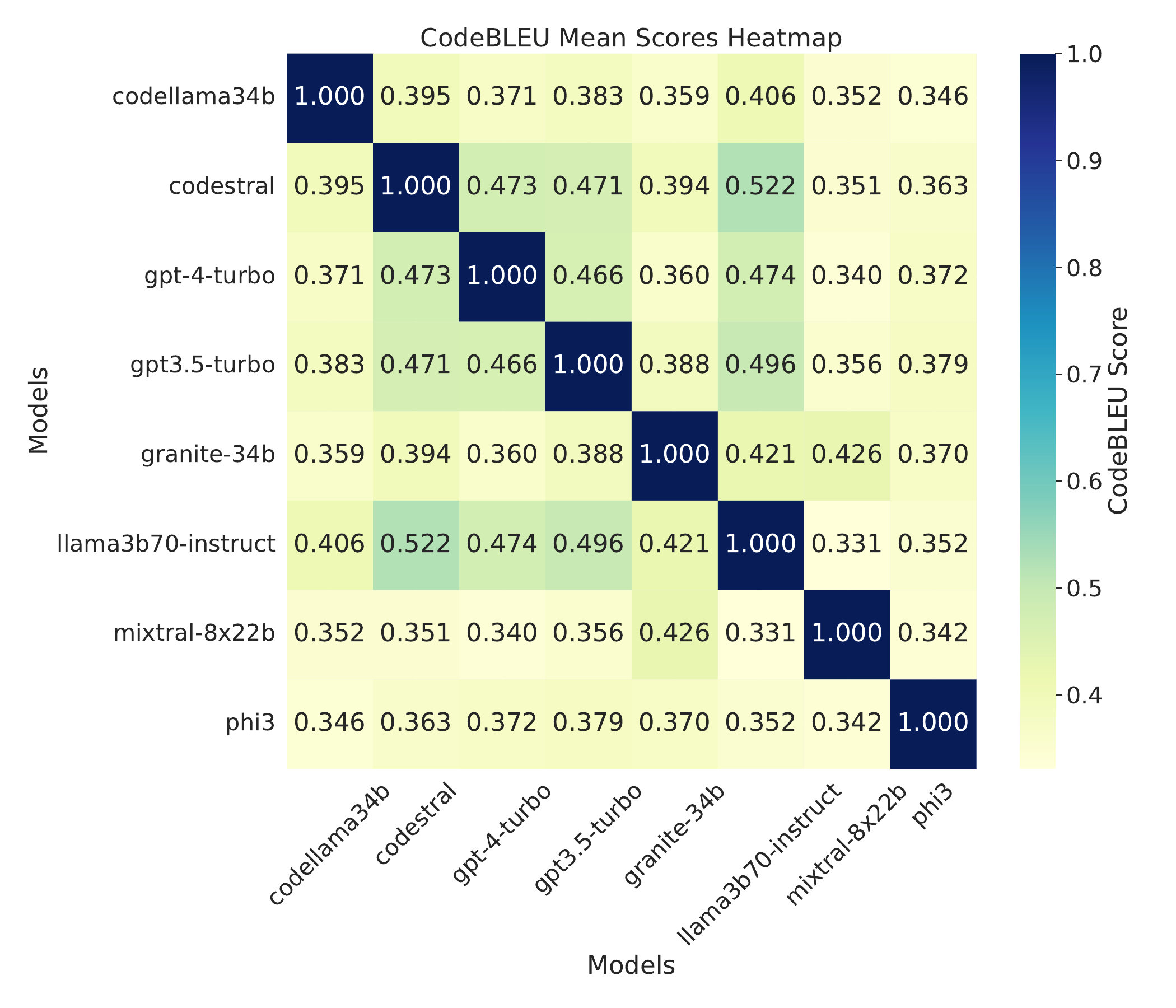}
    \caption{Pairwise zeroshot performance comparison between models for translated Stack-V2 dataset}
    \label{fig:stack_results}
\end{figure}

\subsubsection{Dataset-Specific Performance: HPC Fortran2CPP  vs. Numerical Recipes in Few-Shot setting}

The datasets used in this study, namely HPC Fortran2CPP and Numerical Recipes, exhibited different performance characteristics across the models. The HPC Fortran2CPP dataset generally yielded higher CodeBLEU scores compared to the Numerical Recipes dataset. For instance, Granite-34B Code Instruct with CodeBERT embedding achieved a one-shot CodeBLEU of $0.6$ on the HPC Fortran2CPP dataset, whereas it scored $0.49\pm 0.20$ on the Numerical Recipes dataset. This discrepancy can be attributed to the inherent differences in the complexity and structure of the code in these datasets. The HPC Fortran2CPP dataset, which may contain more standardized and less complex code, allowed models to perform better. In contrast, the Numerical Recipes dataset, with potentially more intricate and varied code structures, posed greater challenges for the models.

\subsubsection{Impact of LLM models}
When comparing different models, Mixtral-8x22B, Llama3-70B, and Granite-34B stood out as the top performers in few-shot settings, whereas GPT-3.5 Turbo and GPT-4 Turbo excelled in zero-shot settings. Starcoder, on the other hand, showed relatively lower performance, particularly in the Zero-shot setting, with a CodeBLEU of 0.21. Additionally, Starcoder did not show significant improvement in one-shot settings, likely due to its smaller context length.

Models such as llama3, Codestral, Mixtral, Granite, and CodeLlama consistently outperformed others across multiple metrics, particularly in terms of CodeBLEU scores. This superior performance can be attributed to their explicit pre-training or fine-tuning for code-related tasks. These models have likely been trained on extensive code-specific data and tasks, enabling them to better understand and generate programming languages. Conversely, models like Phi-3 did not perform as well in comparison, likely because they were not specifically optimized for code-related tasks. While Phi-3 may have been trained on a diverse set of texts, including some programming language data, their training was not as focused on code-specific tasks as Granite and CodeLlama. Consequently, their ability to handle Fortran to C++ translation is less robust.

While GPT models such as GPT-4 Turbo excel in zero-shot settings with high initial CodeBLEU scores, their performance does not increase substantially with additional shots. This plateau in performance can be attributed to GPT models prioritizing the generation of executable and semantically correct code over strict alignment with ground truth translations. As a result, their incremental benefit from additional examples is limited compared to other models.

Starcoder's notably poor performance can be explained by a couple of factors. First, Starcoder is a smaller model with a smaller token limit, which restricts its capacity to understand and generate complex code structures compared to larger models. Additionally, it appears that Starcoder's training may not have been as focused on code-specific data, leading to lower performance even with multiple shots. This suggests that for tasks requiring a deep understanding and generation of code, model size and the specificity of training data play crucial roles in determining performance.

\subsubsection{Results on Unlabelled Dataset in zero-shot settings}
We first translated the StackV2 Fortran dataset using different LLMs and computed the pairwise CodeBLEU similarity between the model translations. The results are presented in Figure \ref{fig:stack_results}. In this experiment, we aimed to evaluate the consistency and similarity of translations produced by various models. The heatmap in Figure \ref{fig:stack_results} illustrates the pairwise CodeBLEU scores, where each cell represents the similarity between translations generated by a pair of models. From the heatmap, we can observe the following key points:  The highest pairwise similarity scores are observed between \textit{Codestral} and \textit{Llama3-70B Instruct} models, with a score of 0.522. The \textit{Granite-34B Code Instruct} model shows relatively high similarity with several models, including \textit{Llama3-70B Instruct} and \textit{Mixtral-8x22B}. Models \textit{CodeLlama-34B Instruct}, \textit{GPT 4 Turbo}, and \textit{GPT 3.5 Turbo} also demonstrate a moderate degree of consistency in their translations. These results indicate that while there is a degree of variability in the translations generated by different models, certain models produce more consistent and similar translations compared to others. The main takeaway from this experiment is that identifying models with reliable and consistent translations guides researchers in model selection for future tasks, thereby enhancing translation accuracy, reliability, and advancing automated code translation methodologies.

\section{Conclusion and Future Work}
In conclusion, the results of this study underscore the importance of model architecture, training data specificity, and the effective use of few-shot learning in enhancing code translation tasks. Models explicitly pre-trained or fine-tuned for code-related tasks, such as GPT-3.5/4 and Llama3-70B-Instruct, demonstrate superior performance, while models with more general training, like Phi-3 3.8B  fall short. The findings also highlight the potential limitations of smaller models like Starcoder in handling complex code translation tasks, emphasizing the need for adequate model size and training data to achieve optimal performance. The use of RAG setups further enhances the translation quality by providing relevant context, proving to be a valuable strategy in few-shot learning scenarios.

The current limitation in Fortran-C++ pairs challenges fine-tuning LLMs and establishing benchmarks. Our Stack-v2 dataset experiments should guide expert translation collections paired with RAG setups, enhancing model performance and applicability across code translation tasks. Future research will expand the dataset and refine the RAG framework to improve translation quality and reliability.

\bibliographystyle{ieeetr}
\bibliography{reference}

\end{document}